\documentclass[letterpaper]{article} % DO NOT CHANGE THIS
\usepackage{aaai25}   % DO NOT CHANGE THIS
\usepackage{times}  % DO NOT CHANGE THIS
\usepackage{helvet}  % DO NOT CHANGE THIS
\usepackage{courier}  % DO NOT CHANGE THIS
\usepackage[hyphens]{url}  % DO NOT CHANGE THIS
\usepackage{graphicx} % DO NOT CHANGE THIS
\usepackage{natbib}  % DO NOT CHANGE THIS AND DO NOT ADD ANY OPTIONS TO IT
\usepackage{caption} % DO NOT CHANGE THIS AND DO NOT ADD ANY OPTIONS TO IT
\usepackage{algorithm}
\usepackage{algorithmic}

\usepackage{newfloat}
\usepackage{listings}

\usepackage{subcaption}
\usepackage[table]{xcolor}
\usepackage{colortbl}

\usepackage{graphicx}
\usepackage{booktabs}       % professional-quality tables
\usepackage{amsfonts}       % blackboard math symbols
\usepackage{nicefrac}       % compact symbols for 1/2, etc.
\usepackage{microtype}      % microtypography
\usepackage{xcolor}         % colors
\usepackage{amsmath}
\usepackage{multirow}
\usepackage{bibentry}
\definecolor{c1}{HTML}{ff0097}

\DeclareCaptionStyle{ruled}{labelfont=normalfont,labelsep=colon,strut=off} % DO NOT CHANGE THIS
\floatstyle{ruled}
\newfloat{listing}{tb}{lst}{}
\floatname{listing}{Listing}
\definecolor{clr_person}{RGB}{192,108,255}
\definecolor{clr_bus}{RGB}{210,103,145}
\title{InvSeg: Test-Time Prompt Inversion for Semantic Segmentation }

\author{
    Jiayi Lin\textsuperscript{\rm 1}, 
    Jiabo Huang\textsuperscript{\rm 2}, 
    Jian Hu\textsuperscript{\rm 1}, 
    Shaogang Gong\textsuperscript{\rm 1}
}
\affiliations{
    \textsuperscript{\rm 1}Queen Mary University of London\\
     \textsuperscript{\rm 2}Sony AI\\
    \large{\textcolor{c1}{https://jylin8100.github.io/InvSegProject/}}
}

\usepackage{bibentry}
\begin{document}
\maketitle

\begin{abstract}
Visual-textual correlations in the attention maps derived from text-to-image diffusion models are proven beneficial to dense visual prediction tasks, e.g., semantic segmentation. However, a significant challenge arises due to the input distributional discrepancy between the context-rich sentences used for image generation and the isolated class names typically used in semantic segmentation. This discrepancy hinders diffusion models from capturing accurate visual-textual correlations. To solve this, we propose InvSeg, a test-time prompt inversion method that tackles open-vocabulary semantic segmentation by inverting image-specific visual context into text prompt embedding space, leveraging structure information derived from the diffusion model's reconstruction process to enrich text prompts so as to associate each class with a structure-consistent mask. Specifically, we introduce Contrastive Soft Clustering (CSC) to align derived masks with the image's structure information, softly selecting anchors for each class and calculating weighted distances to push inner-class pixels closer while separating inter-class pixels, thereby ensuring mask distinction and internal consistency. By incorporating sample-specific context, InvSeg learns context-rich text prompts in embedding space and achieves accurate semantic alignment across modalities. Experiments show that InvSeg achieves state-of-the-art performance on the PASCAL VOC, PASCAL Context and COCO Object datasets.

\end{abstract}

\section{Introduction}

Open-Vocabulary Semantic Segmentation (OVSS) aims to divide an image into several semantically consistent regions, corresponding to label names in a large, unrestricted vocabulary. 
Recently, thanks to visual-textual correlation ability learned from large scale image-text pairs, stable diffusion models~\cite{Stable_diffusion} have shown substantial potential in open vocabulary semantic segmentation.
Some approaches employ the stable diffusion model for generating pseudo labels~\cite{wu2023diffumask,wang2024image, li2023open, xiao2023text} 
or use it as a feature extractor for further training a segmenter~\cite{odise}.
However, among these approaches, the process of collecting per-pixel labels is costly and 
 the performance can be limited when generalizing to new scenarios under distribution shifts. 
Therefore, to eliminate the need for expensive pixel-level annotation and achieve better generalization ability, several unsupervised approaches~\cite{diffseg, ovdiff, diffsegmenter} are proposed and have achieved notable results.

\setlength{\fboxsep}{1pt}
\begin{figure}[tb!]
\centering
    \begin{subfigure}[b]{0.45\textwidth}
        \centering
        \includegraphics[width=\textwidth]{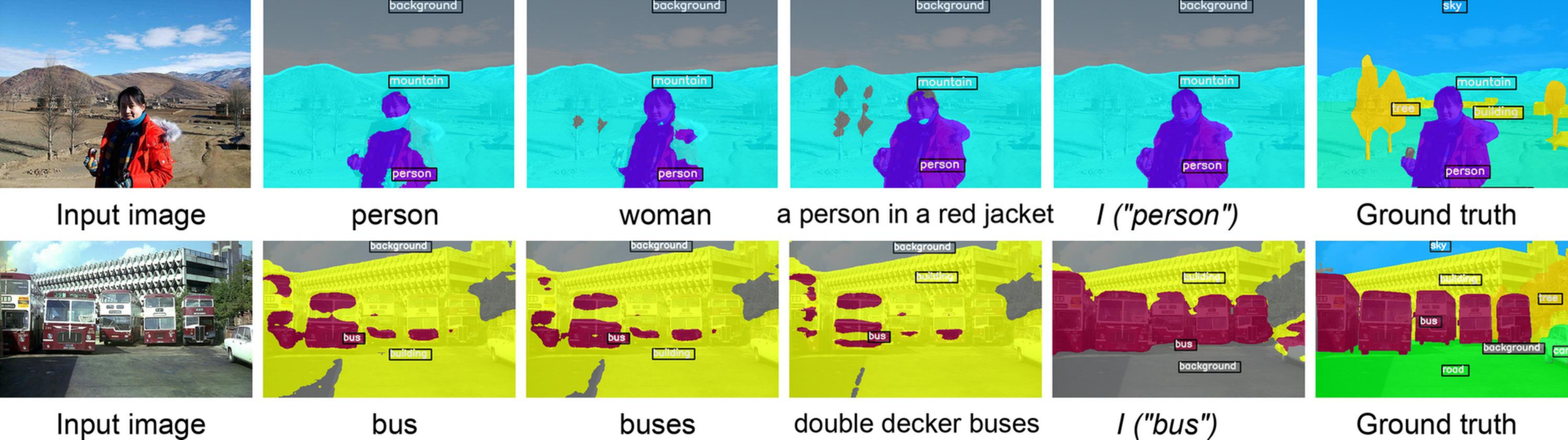} 
        \caption{Comparison of segmentation results for  ``\colorbox{clr_person}{person}'' or ``\colorbox{clr_bus}{bus}'' using text prompts with different levels of image-specific details.}\label{fig1:a}
        \label{fig:sub2}
    \end{subfigure}
    \begin{subfigure}[b]{0.45\textwidth}
        \centering
        \includegraphics[width=\textwidth]{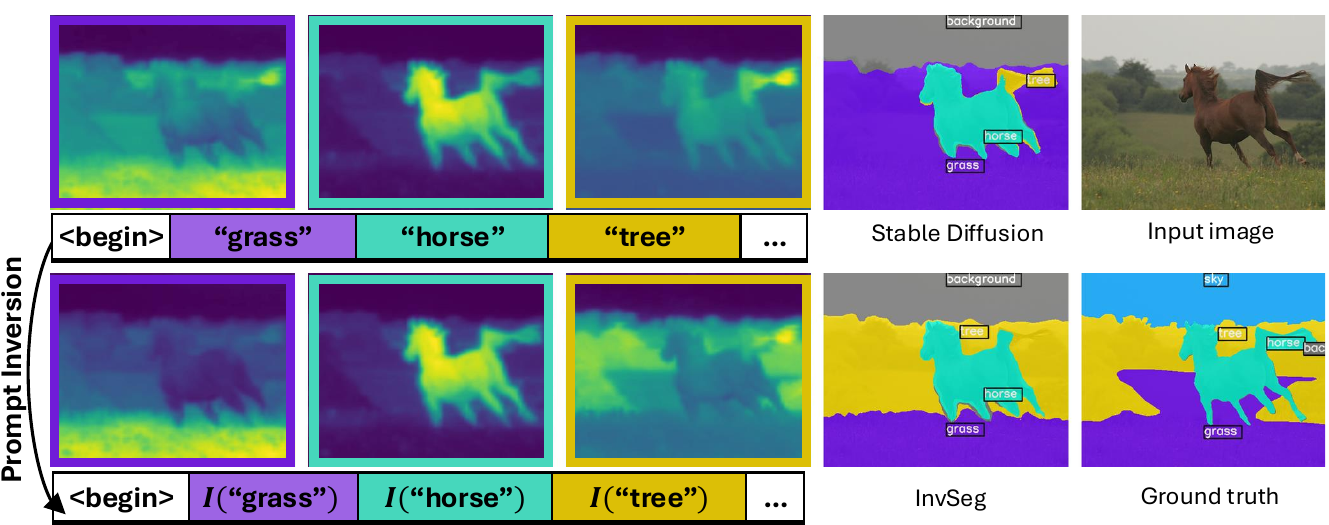} 
        \caption{Overall process of InvSeg. 
  We 
  exploit the structural information from image to enrich the original isolated class names (top) 
  into image-specific text prompt which
  derives more accurate segmentation results (bottom).}\label{fig1:b}
    \end{subfigure}
  \caption{Motivation of test-time prompt inversion. 
  }
  \label{fig:loss_para}
\end{figure}

% (已有的方法没有很好解决）
However, most current unsupervised approaches~\cite{diffseg, ovdiff}
    solely use isolated class names
    as the text prompt for diffusion models,
    which ignores the distributional discrepancy 
    between the isolated class names for visual perception and 
    descriptive, context-rich sentences used for image generation.
This discrepancy in input text richness 
    can hinder the performance of diffusion models
    in semantic segmentation,  
    where precise image-text correlations is required.
Although some approaches~\cite{diffsegmenter} enrich the class name 
    by combining it with other descriptive text of current class,
this process is cumbersome and 
    may not always derive to satisfying output.
    % as the 
    % description may not contain enough the information for segmentation.
As in our preliminary study (Fig.~\ref{fig1:a}),
    we evaluate text prompts with varying levels of
    sample-specific information: 
    original class name, 
    sample-specific single word, 
    sample-specific word group,
    where the latter two are extracted from the captions generated by a vision language model (VLM)~\cite{li2022blip}. 
We observe that containing more sample-specific information can enhance segmentation performance. 
However, in the second example, we observe that increasing sample-specific information does not necessarily improve segmentation results. 
This inconsistency shows the complexity of 
    searching optimal text prompts for segmentation,
    which requires searching a carefully crafted combination of different words.

To avoid the complexity of
    searching multiple words for segmenting single class, 
    we propose to optimize the original class name 
    in text embedding space by
    extracting sample-specific visual context.
We propose InvSeg,
    a test-time prompt inversion method,
    which inverts 
    a single test image 
    into the model's text prompt embedding space, where each text token captures a distinct visual concept in the image.
Specifically, 
    we introduce Contrastive Soft Clustering (CSC) 
    to align the predicted masks
    with inherent image structure.
CSC constructs a 4D distance matrix 
    that measures pairwise distances between spatial points in an image, 
    with each spatial point having its corresponding distance map 
    illustrating its distance to other points.
In each predicted score map corresponding a specific class,
    some high-confidence points are selected as the anchor points and their distance maps is used to aligned with current predicted mask.
This alignment process 
    involves utilizing each anchor point's distance map 
    to modulate the probability distribution across classes. 
Particularly, we leverage the distance information to determine the extent 
    to which we decrease the probability of the current class
    and increase the probabilities of other classes for each spatial point.
By incorporating sample-specific context at test time, InvSeg 
learns a context-rich text prompt
to achieve precise semantic alignment across modalities.

We benchmark InvSeg on three segmentation datasets PASCAL VOC 2012~\cite{everingham2010pascal}, PASCAL Context~\cite{pascal0context} and COCO Object~\cite{coco} for OVSS task.
InvSeg outperforms prior works on both datasets despite requiring less auxiliary information. In summary, our contributions are two-fold:
\begin{itemize}
% \item We uncover the necessity of customizing/optimize the image-specific text prompts 
%     to boost the potential of generative text-to-image diffusion models to locate more diverse visual concepts.

\item To the best of our knowledge, we are the first to perform region-level prompt inversion on diffusion models in an unsupervised manner. This enables the decomposition of any image into semantic parts for visual perception tasks and serves as a foundation for visual reasoning tasks.

\item With our proposed unsupervised constraint, we are able to obtained optimized image-specific text prompt to generate the more complete and accurate attention maps and derive superior segmentation results comparing to previous unsuperivsed methods. 
\end{itemize}

% （logic 也是要根据abstract）
\section{Related Work}
% 主次，三段的关系？2，3包含在1里面吗？
% 【【可以解决这个问题的方法有什么类别，】
% 为什么一个section有这几段，这一段里有这几句。
% [1] Open-vocabulary Object Segmentation with Diffusion Models. In ICCV, 2023.
% [2] DatasetDM: Synthesizing Data with Perception Annotations Using Diffusion Models. In NeurIPS, 2023.
% [3] Diffusion Models for Zero-Shot Open-Vocabulary Segmentation. In arXiv 2023.
% [4] MosaicFusion: Diffusion Models as Data Augmenters for Large Vocabulary Instance Segmentation. In arXiv, 2023.
\subsection{Pre-trained Generative Models for Segmentation.}
Pre-trained Generative Models like Generative Adversarial Nets~\cite{goodfellow2014generative, tritrong2021repurposing, xu2021linear} and Stable Diffusion Models~\cite{Stable_diffusion} have been widely used in dense prediction like segmentation for its ability to capture fine-grained location  and shape information of objects in images. 
Recently, there are increasing interest in exploiting diffusion models
for segmentation~\cite{wu2023diffumask,wang2024image, li2023open, xiao2023text,wu2023diffumask}. Some approaches use the model internal features to train a segmentation branch, other use  diffusion models to generate new segmentation data to train a segmenter~\cite{wu2023diffumask}. 
However, these models  require expensive per-pixel
labels or labeling process, which  tend to suffer from the category imbalance problem.
In contrast, current unsupervised methods
use attention maps from pre-trained denoising U-Nets in diffusion models.
% where the cross-attention maps contain semantic information  and the self-attention maps contain pairwise pixel affinities that shows object shape~\cite{diffseg, diffsegmenter}.
Specifically, DiffSeg~\cite{diffseg}  uses self-attention maps to get pairwise pixel similarity  to cluster pixels iteratively, which however the output mask does not align to semantic category, and
requires manual design of the number of clusters.
DiffSegmenter~\cite{diffsegmenter} uses the cross-attention maps to initial the mask and self-attention maps to complete the mask for each given category. 
However, the text embedding of the same category in these methods remain the same for different images, while we instead propose to customize the text embedding to get more complete and accurate segmentation masks for each image.

\begin{figure*}
    \includegraphics[width=0.99\textwidth]{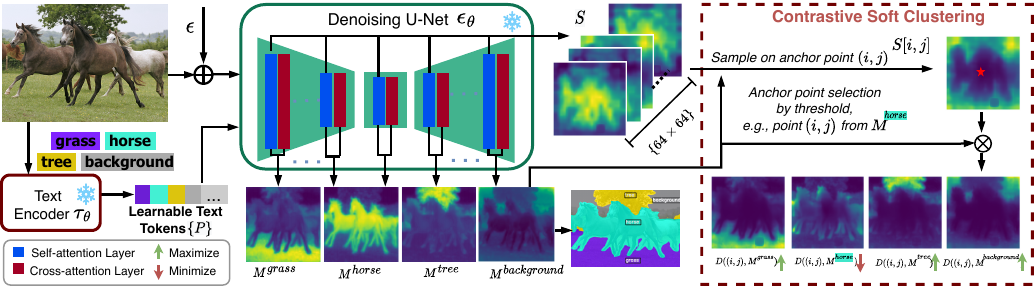}
  \caption{Overview of InvSeg framework. Our proposed Contrastive Soft Clustering method can achieve region-level prompt inversion. The text tokens are first initialized with the pretrained text encoder from the diffusion model (dashed box on left) and then are used as the only learnable parameters during the test time training. After the adaption process, the learned text tokens can be used to derive a more accurate and complete refined attention maps $\{M\}$ for segmentation.
   }
\end{figure*}

% 【主次，方法实现的而已，有多重要】，是paragram还是subsection？
\subsection{Diffusion-based Inversion. }
Diffusion-based Inversion is inspired from GAN inversion~\cite{bojanowski2017optimizing}
to synthesize visual concept in the given reference images for further image editing.
Specifically, DreamBooth~\cite{ruiz2023dreambooth} fine-tunes the diffusion model to learn to bind a unique identifier with that specific subject.
In contrast, some other inversion approachs~\cite{mokady2023null, gal2022image} keep the model weights  fixed to avoid damaging the knowledge of the pre-trained model  and cumbersome
tuning process.
Textual Inversion~\cite{gal2022image} inverts visual concepts in the text prompt embedding space and encode the visual features as new tokens in the vocabulary. 
Similar in spirit, we introduce to invert new tokens for regions in an image for category segmentation. The only work that also try to learn region level prompt is SLIME~\cite{khani2023slime}, which uses mask annotation of few sample  to learn region-level prompt and transfer to similar images. However, this method needs annotation data and requires the test images to be similar to the training image, which greatly inhibits the model's transferability to broader diverse images. Therefore, the method is only tested on a few part segmentation datasets like horse, face and car part segmentation. In contrast, InvSeg can be applied for more diverse images in an unsupervised way.

% 【为什么当下的tta方法不能解决，不trival； 】
\subsection{Test Time Prompt Learning. }
Test Time Prompt Learning is a subtopic of Test Time Adaptation~\cite{sun2020test}, which narrows the distribution gap between the
training and test data during test time.
Test-time Prompt Tuning (TPT)~\cite{shu2022test} learns adaptive prompts of a large VLM CLIP~\cite{clip} for one test sample, by forcing consistent predictions across different augmented views of each test sample with an entropy minimization
objective.
PromptAlign~\cite{abdul2024align} further introduce a distribution alignment objective for CLIP and explicitly aligns the train and test sample distributions to mitigate the domain shift.
We exploit the entropy minimization constraint during the optimization of text prompt, to stabilize the adaptation process.

% （没有formulate 好； 没有recap整个story？ ）
\section{Methodology}
InvSeg aims at inverting the image-specific fine-grained text prompts from a given test image using pretrained diffusion model, where the inverted prompts are further used to produce high quality segmentation masks.
We first exploit the original text prompts (category names) to generate segmentation masks with diffusion models,
which provide initial segmentation masks for further adaptation. % (Sec.~\ref{sec:dm}).
Then 
we introduce use Contrastive Soft Clustering to achieve region-level prompt inversion in an unsupervised way, which helps segment complete and disjoint masks. % (Sec. \ref{sec:csc}).
Finally, we stabilize the adaption process  with Entropy Minimization. %(Sec. \ref{sec:stabilize}).

Given a test image $I \in  \mathbb{R}^{h \times  w \times  3}$ and a list of categories from the test dataset. We first use Vision Language Model~\cite{li2022blip} to filter out the $C$ categories (including one  category as "background") in current image following previous works~\cite{diffseg, diffsegmenter}. 
Then the filtered category are directly combined as the  text prompt input.

\subsection{Diffusion Models}\label{sec:dm}
The text-conditioned diffusion model models a data distribution conditioned a natural language text prompt. The text prompt $y$ is first tokenized and then encoded by a text encoder $\tau_\theta$ to obtain text token embeddings $\tau_\theta(y)=\{P^k_0\}$, where k is the index of each token. 
On the other hand, the image $I \in  \mathbb{R}^{h \times  w \times  3}$ is first encoded to auto-encoder latent space, and then added with a standard Gaussian noise $\epsilon$ for  $t$ 
 time step to obtain $\mathcal{I}_t$.
Finally, the objective is to predict the added noise with diffusion model noted as $\epsilon_\theta$, where the text encoder $\tau_\theta$ and diffusion model $\epsilon_\theta$ are optimized simultaneously:
\begin{equation}
L_{L D M}=\mathbb{E}_{\mathcal{I}, y, \epsilon \sim \mathcal{N}(0,1), t}\left[\left\|\epsilon-\epsilon_\theta\left(\mathcal{I}_t, t, \tau_\theta(y)\right)\right\|_2^2\right]
\end{equation}
In InvSeg, we keep the parameters of the models  $\tau_\theta$, $\epsilon_\theta$ fixed, but optimize $\{P^k_0\}$ directly to obtain an image-specific prompt $\{P^k_I\}$  for image $I$.
% Following prior prompt inversion methods~\cite{gal2022image}, we use $L_{L D M}$ as the loss function for image-level inversion during adaption.

\noindent\textbf{Attention Map Generation in Diffusion Models.}
Diffusion model employs a UNet structure of four resolutions $\{8,16,32,64\}$, where the attention modules~\cite{vaswani2017attention} are applied on each resolution for multiple times.
These attention modules include self-attention and cross-attention. 
Self-attention 
captures pixel-level affinities within the image, while cross-attention captures the relationship between the text tokens and image pixels in embedding space.
For each attention module, there are three components: query $Q$, key $K$, and value $V$ of dimension $d$. The output of attention is $O=\operatorname{Softmax}\left(\frac{Q K^{\top}}{\sqrt{d}}\right) \cdot V$.

Here we use a normalized attention map
$S=\operatorname{Softmax}\left(\frac{Q K^{\top}}{\sqrt{d}}\right)$ for further computation following previous work~\cite{diffsegmenter},
which shows the similarity between  query $Q$ and key $K$.
Therefore, 
in a self-attention module, we obtain $S^{self}$ that capture pixel-level similarities in the image embedding space. 
Similarly, in a cross-attention module, $S^{corss}$ is obtained to measure the similarities between each text token and image pixels.
\begin{figure}[b]
    \includegraphics[width=0.47
    \textwidth]{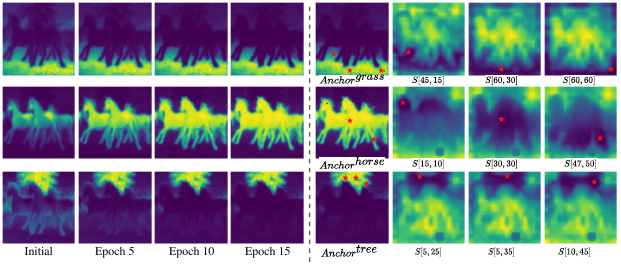}
    \vspace{-10pt}
  \caption{Illustration of the soft selection (with probability) of anchor points for each category $c$: $Anchor^c$ during different optimization steps (left) and the distance matrix $S$ on certain anchor points (right).  
On the right sub-figure, we sample 3 anchor points for each category, showing the distance from each anchor point to other pixels in the image. Darker areas represent smaller distances (higher similarity) to the anchor. }
\end{figure}
We further aggregate the all the $\{S^{self}_l\}$ from different self-attention modules $l$ respectively for further computation, which is also performed on $\{S^{cross}_l\}$  following~\cite{vaswani2017attention}.
Taking  $\{S^{self}_l\}$  as an example, 
we first average those with the same resolution and interpolate them to the resolution of $64\times64$. 
Then we use a weighted sum among all the resolutions to obtain a final self-attention map 
$\mathcal{A}^{ {self}} \in \mathbb{R}^{H  W \times H  W}$, where $H=W=64$. 
In a similar way, we obtain $\mathcal{A}^{{cross}} \in \mathbb{R}^{H  W \times K} $, where K is the number of tokens. 
Finally, we use a mask refinement mechanism in  DiffSegmenter~\cite{vaswani2017attention} which
uses $\{S^{self}_l\}$ to incomplete $\mathcal{A}^{{cross}}$ to get a refined cross-attention map:
\begin{equation}
M=\operatorname{norm}\left(\mathcal{A}^{{self}} \cdot \mathcal{A}^{{cross}}\right),
\end{equation}
where $\operatorname{norm}(\cdot)$ is min-max normalization.
Therefore, for each token k, we have a corresponding refined cross-attention map $M^k$. 
Furthermore, for each category, we use notation $c$ as the index. Therefore, we can further obtain a set of refined cross-attention map ${M^c}$ from ${M^k}$.

\subsection{Contrastive Soft Clustering}\label{sec:csc}
% $\mathbb{Z}$
As the refined cross-attention map ${M^c}$ derived directly from initial text token embeddings ${P^c_0}$ can only locate part of the object (Fig.~\ref{fig1:b}) and tend to be distracted by salient category when locating background category (Fig.~\ref{fig1:a}), which leads to unsatisfying segmentation results.
We propose Contrastive Soft Clustering to achieve region-level inversion for each text token (category) to  generate complete and disjoint attention map for each category by exploiting the structure information in the pretrained diffusion models.

Specifically, after the refined cross-attention maps {$M^c$} are obtained, we further optimize {$P^c$} by ensuring that its corresponding {$M^c$} is aligned with the objectness information indicating in self-attention maps.
As illustrated in previous works~\cite{diffseg, diffsegmenter}, the self-attention maps 
contain specific structural information of object shape and location. 
Based on this observation, we build a 4D distance matrix $S$ to measure the distance between each pair of pixels in the refined cross-attention map $M^c$.
We measure the distance between two attention maps using KL divergence following~\cite{diffseg}. Therefore, the distance between pixel $(i,j)$ and  pixel $(k,l)$ is:
% \begin{equation}
\begin{multline}
\mathcal{S}[i,j,k,l] =\operatorname{KL}\left(\mathcal{A}^{\text {self}}[i, j] \| \mathcal{A}^{\text {self}}[k, l] \right) \\
+\operatorname{KL}\left(\mathcal{A}^{\text {self}}[k, l] \| \mathcal{A}^{\text {self}}[i, j] \right)
\end{multline}
% \end{equation}

Based on the distance matrix $S$ and $M^c$,
we can measure the weighted distance between pixel $(i,j)$ and a group of weighted pixels in $M^c$ as follows:
\begin{equation}
D((i,j),M^c)=\frac{\sum_{(k,l)\in Q^c}\left(\mathcal{S}[i,j,k,l] \cdot M^c[k,l] \right)}{\sum_{(k,l)\in Q^c}M^c[k,l]}
\end{equation}
where $Q^c$ is the set of all the pixels in $M^c$.

To measure the distance of all the pixels in $M^c$, we use sigmoid to binarize the value in $M^c$, which gets a soft selection of high confident pixels corresponding to the current category $c$.
We denote this soft selected set as $Anchor^c\in \mathbb{R}^{H \times W}$, where the pixels with value close to $1$ is used as an anchor of category $c$. 
Therefore, the distance within $M^c$:
\begin{multline}
% \begin{multline}
D(Anchor^c,M^c)= \\
\frac{\sum_{(i,j)\in Q^c}\left( Anchor^c[i,j] \cdot D((i,j),M^c) \right)}{\sum_{(i,j)\in Q^c} Anchor^c[i,j]}.
\end{multline}
Similarly, we can easily calculate the distance between different categories $c$ and  $c^{\prime}$, that is $D(Anchor^c,M^{c^{\prime}})$.
% \begin{equation}
% D(Anchor^c,M^{c^{\prime}})=\frac{\sum_{(i,j)\in Q^c}\left( Anchor^c[i,j] \cdot D((i,j),M^{c^{\prime}}) \right)}{\sum_{(i,j)\in Q^c} Anchor^c[i,j]}.
% \end{equation}
Therefore, for all $C$ classes, the total distance within each class is:
\begin{equation}
D_{intra}=\sum^C_{c=1}D(Anchor^c,M^c)
\end{equation}
and the total distance of all pairs of two different classes is:
\begin{equation}
D_{inter}=\sum^{C-1}_{c^{\prime}=1}\sum^C_{c=c^{\prime}+1}D(Anchor^c,M^{c^{\prime}})
\end{equation}
In this way, we are able to perform a contrastive soft clustering by minimizing the distance within each category and maximizing the distance among different categories.
The loss function of contrastive soft clustering is denoted as: 
\begin{equation}
\mathcal{L}_{Cluster}=\frac{D_{intra}}{C}-\frac{2*D_{inter}}{C*(C-1)}
\end{equation}

% sota
\begin{table*}[ht]
\caption{\textbf{Comparison with existing methods.} Models in the first three rows are finetuned on target datasets while the rest approaches do not require mask annotations. \textbf{Bold fonts} refer to the best results  and \underline{underline fonts} refer to the second best.}\label{tab:sota}
\centering
\vspace{-1pt}
% \vspace{5pt}
\resizebox{0.99\textwidth}{!}{
  \begin{tabular}{lcccc}
    \toprule
    {\multirow{2}{*}{\centering{Methods}}} & 
    % {\multirow{2}{*}{\centering{Architect}}} &
    {\multirow{2}{*}{\centering{Training dataset}}} & 
\multicolumn{3}{c}{mIOU}
\\ \cline{3-5}  
     & & 
      {\centering{PASCAL VOC}} & 
      PASCAL Context & 
      COCO Object \\
    \midrule
    \midrule
    
    DeiT \cite{deit}  & IN-1K   &  53.0 & 35.9 & - \\
    MoCo \cite{moco}  & IN-1K  & 34.3 & 21.3 & - \\
    DINO \cite{dino}  & IN-1K & 39.1 & 20.4 & - \\

    \midrule
    ViL-Seg \cite{vilseg}  & CC12M &  33.6 & 15.9 & - \\

    MaskCLIP~\cite{maskclip}         & LAION    & 38.8 & 23.6 & 20.6 \\
    GroupViT~\cite{xu2022groupvit}   & CC12M   &  52.3 & 22.4 & - \\
    ZeroSeg~\cite{zeroseg}            & IN-1K  & 40.8  & 20.4 & 20.2 \\
    TCL~\cite{tcl}              & CC3M+CC12M       &  51.2 & 24.3 & 30.4 \\
    ViewCo~\cite{ren2023viewco}  &CC12M+YFCC &52.4 &23.0 &23.5 \\
    CLIPpy \cite{clippy}  & HQITP-134M &  52.2 & - & 32.0 \\
    SegCLIP~\cite{luo2023segclip}   & CC3M+COCO & 52.6 & 24.7 & 26.5 \\
    OVSegmentor~\cite{ovsegmenter} & CC4M &  53.8 & 20.4 & 25.1 \\
    SimSeg~\cite{yi2023simple}  & CC3M+CC12M        &  57.4 & 26.2 & 29.7 \\
    \midrule
    \emph{Generative models based} \\
    % \emph{Sequential Binary Segmentation  w/o Caption} \\
     % Diffusion Baseline~\cite{Stable_diffusion} &UNet+ViT& -  &  \\  
     % InvSeg (Ours)  & UNet+ViT& -  &\\
    % \midrule
    % \emph{Sequential Binary Segmentation w/ Caption} \\
    % Diffusion Baseline(sd1.5)~\cite{Stable_diffusion}  & UNet+ViT & -  & 59.5  & 25.4 & 34.6 \\ 
    % Diffusion Baseline~\cite{Stable_diffusion}  & UNet+ViT & -  & 57.9  & 24.1 & 33.0 \\ 
    DiffSegmenter~\cite{diffsegmenter} & -  & \underline{{60.1}}  & \underline{{27.5}} & \textbf{37.9} \\
    % InvSeg(sd1.5) (Ours) & UNet+ViT        & -  & & & \\ 
    % InvSeg (Ours) & UNet+ViT        & -  &59.9 & 25.9 & 33.3 \\
    % \midrule

    % \emph{Simultaneous Multi-Category  Segmentation w/o Caption} \\
    % OVDiff~\cite{ovdiff} & UNet+ViT  & - &  \textbf{67.1}   & \textbf{30.1} & 34.8 \\     % by finding a visual fg & bg prototype, n match with an image in pixel level? how to get the class name?
        % Diffusion Baseline(sd1.5)~\cite{Stable_diffusion}& UNet+ViT   & -  & {57.1} & {24.4}& {33.8} \\
    % InvSeg(sd1.5) (Ours) & UNet+ViT        & -  & {57.2}& {25.5} & {32.3} \\ 
    Diffusion Baseline~\cite{Stable_diffusion}   & -  & 59.6 & 25.0 & 34.5  \\

    % InvSeg (Ours)         & -  & & &  \\
    InvSeg (Ours)         & -  & \textbf{63.4}& \textbf{27.8} & \underline{36.0} \\
    % Diffusion Baseline* & UNet  & - & 59.3  &31.8 &29.7  \\
    % InvSeg* (Ours) & UNet   & - & 64.4  & 33.5& 31.8 \\
    \bottomrule
  \end{tabular}}
\end{table*}
% sota
\vspace{10pt}
\begin{figure*}[ht!]
\centering
 \includegraphics[width=\textwidth]{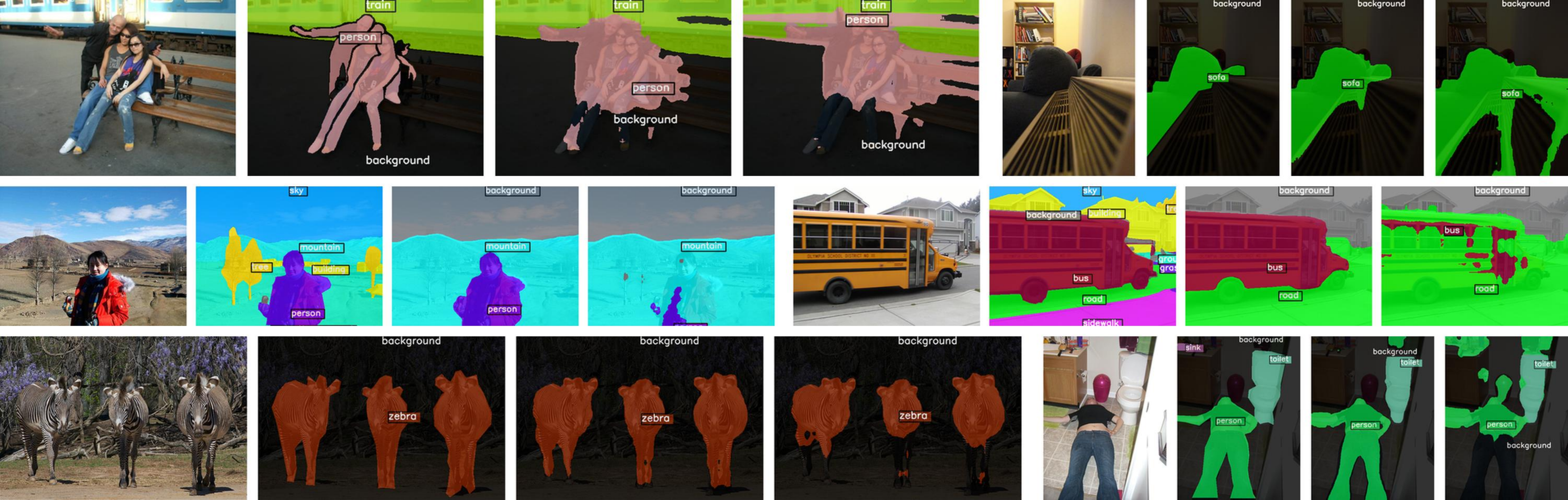}  
  \caption{Examples of Segmentation on VOC (top), Context (middle) and COCO (bottom). For each sample (image group of four), from left to right is input, GT, InvSeg, Diffusion baseline.  }
  \label{fig:sota1}
\end{figure*}

\subsection{Stablizing Adaptation Process}\label{sec:stabilize}
% During the adaptation process with both image-level and region inversion, each text token embedding may drift from original region in the image to other regions, since the one-to-one correspondance of the token and region is not ensured. Therefore, we use the following two constraints to ensure the learnable text token embedding maintain the same semantic information.
\textbf{Entropy Minimization.} Inspired by TPT~\cite{shu2022test}, we augment the input image into different views, and try to encourage consistent predictions across views. To do this, we use   different augmentation functions ${Aug_i}$ on the input image, and get the corresponding attention maps $M_{Aug_i}^c$. Then we reverse the augmentation of align pixels between different maps $M_{Aug_i}^c$ to get $M{\prime}_{Aug_i}^c$. 
Further, we
average the $M{\prime}_{Aug_i}^c$ over different augmented view to get $\overline{M}^c$.
Finally, we minimize the entropy of $\overline{M}^c$ for all categories:
$\mathcal{L}_{Etrp} =-\sum^C_{c=1} \overline{M}^c \cdot\log\overline{M}^c$.

\noindent\textbf{Overall Optimization Process.} For each input image, we optimize the text embeddings with 15 steps, using a loss function $\mathcal{L}=\mathcal{L}_{Cluster} +\alpha*\mathcal{L}_{Etrp}$, where $\alpha$ is the coefficient of the loss functions.
 Finally, we generate a segmentation mask by performing an argmax $M^c$ across differnet categories $c$ and interpolate it to match the original size of input image.
 
\begin{figure*}[ht]
\centering
 \includegraphics[width=\textwidth]{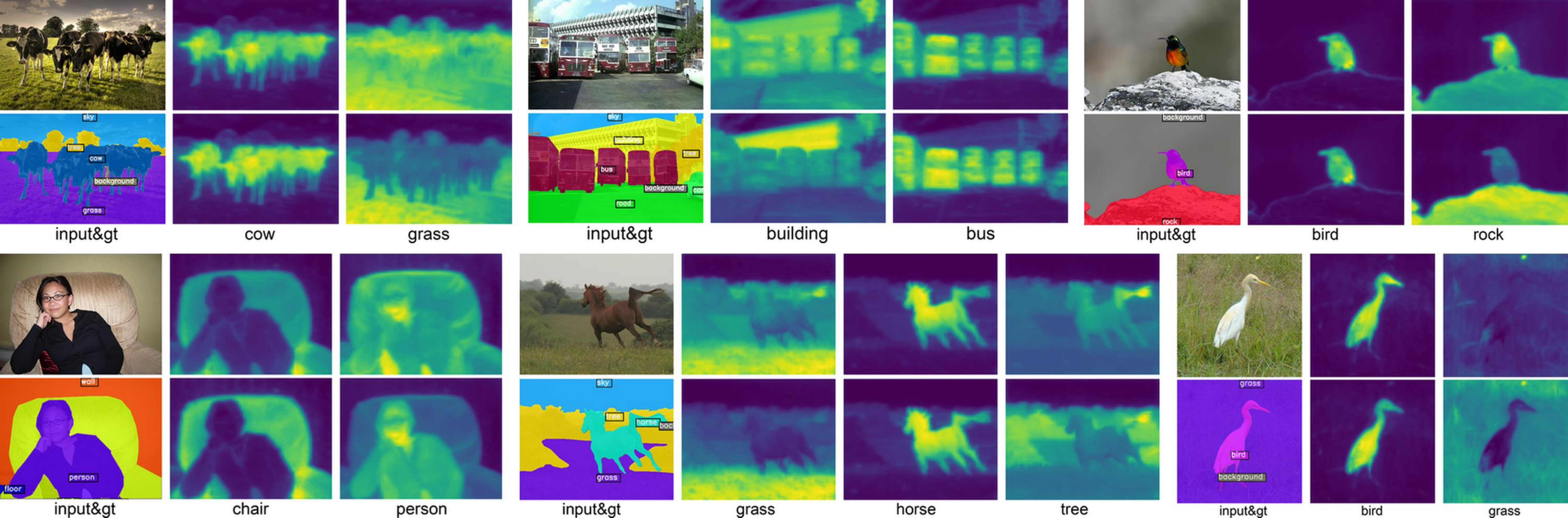} 
\vspace{-8pt}
  \caption{Visualization of refined cross-attention maps derived from text prompts before (top) and after (bottom) prompt inversion. Before prompt inversion, the segmentation of background elements such as "grass" or "trees" is influenced by foreground objects like "cow" or "horse", resulting in mistakenly ignoring background classes or  segmenting foreground (and background) classes. After applying prompt inversion, this phenomenon is suppressed by improving the distinction between foreground and background through proposed Contrastive Soft Clustering.  }
  \label{fig:baseline}
\end{figure*}

\section{Experiments}

\textbf{Datasets and Metrics.} We evaluate InvSeg on three commonly used benchmarks, namely, PASCAL VOC 2012~\cite{everingham2010pascal}, PASCAL Context~\cite{pascal0context} and COCO Object~\cite{coco}, containing 20,59,80 foreground classes, respectively. 
The experiments are performed on the validation sets, including 1449, 5105, and 5000 images.
Following prior works~\cite{diffseg, ovdiff, diffsegmenter}, we use mean intersection over union (mIoU) to measure segmentation performance. We also use the metric mean accuracy (mAcc) in our ablation study.

\noindent\textbf{Implementation Details.}
To obtain  category names, we follow DiffSegmenter~\cite{diffsegmenter} that use BLIP~\cite{li2022blip} and CLIP~\cite{clip} to generate the category names out of all the candidate categories in the dataset. 
We validate our approach using Stable Diffusion v2.1~\cite{Stable_diffusion} with frozen pre-trained parameters.
Each test image is augmented for 2 times using random resized crops with minimum crop rate 0.6, which construct a batch of size 2 in each optimization step. 
We employ the Adam optimizer with a learning rate of 0.01.
The weights for $\mathcal{L}_{Etrp}$ is $\alpha=1$.
We optimize each image for 15 steps on single H100 GPU, with the inference time of around 7.9 seconds per image (not including the category name extraction), 
which comparative to existing test-time prompt tuning methods like our adapted TPT~\cite{shu2022test} for segmentation with 7.2 seconds for inference. 
The running memory was 32.4G on GPU with full precision (32-bit floating point).
% 
 % As for time step, during adaption, we randomly sample from step 5 to 300 following previous work and use step=50 during inference.
% 
As for time step in diffusion model, during adaption, the time step for each iteration is sampled from a range 
 [5, 300] where the model can learn a more robust prompt from different time steps. While the time step for inference is 50, falling in the  range during adaption.
 % , as we only conduct one pass through the diffusion model using the learned prompt.

% 利用stable diffusion的物体概念，将输入的抽象的query，根据图片的信息具象化（离散的token具象化困难）。

\subsection{Results on Open-vocabulary Semantic Segmentation}
% Tab. \ref{tab:sota} shows a comparison of InvSeg to previous works on open-vocabulary semantic segmentation. 
% Previous works mainly include discriminative model (CLIP~\cite{clip}) based methods as well as generative model
Tab. \ref{tab:sota} shows a comparison between InvSeg and previous works in open-vocabulary semantic segmentation. These prior approaches can be broadly categorized into two main groups: those based on discriminative models (such as CLIP~\cite{clip}) and those based on generative models (stable diffusion models~\cite{Stable_diffusion}).
Note that a direct comparison with most diffusion model-based methods would not be equitable. This is because many of these methods use extra mask annotations or synthesis mask which usually requires a pretrained segmenter, such as
generating pseudo labels~\cite{wu2023diffumask,wang2024image, li2023open, xiao2023text, marcos2024open}
or using diffusion models as a backbone to train a model in a supervised manner~\cite{odise,zhao2023unleashing}.
Therefore, we have focused our comparison on a select few diffusion model-based methods that, do not require mask annotations:
DiffSegmenter~\cite{diffsegmenter},
the original Stable Diffusion model (referred to as Diffusion Baseline).
We observe that InvSeg achieves competitive performance compared to previous CLIP based models. 
InvSeg demonstrates state-of-the-art performance on both VOC and Context datasets. When compared to the diffusion baseline that utilizes the original text prompt, InvSeg shows significant improvements in mIOU, with gains of up to 3.8\% on VOC and 2.8\% on Context.
However, InvSeg still lags behind DiffSegmenter on the COCO dataset. This discrepancy may be caused by the fundamental difference in the segmentation strateg. DiffSegmenter employs a sequential strategy, predicting masks for each class individually through binary segmentation. In contrast, InvSeg efficiently segments all classes simultaneously, which is a more complex challenge. 
% This difference in methodology could explain the performance gap observed on the more intricate COCO dataset.
% \subsection{Ablation Studies and Analyses}
\begin{table}[t]
\caption{Ablating loss weights for $\mathcal{L}_{Cluster}$ and $\mathcal{L}_{Etrp}$.}\label{tab:abla_etrp}
\centering
\vspace{-8pt}
\setlength{\tabcolsep}{10pt}
\resizebox{0.48\textwidth}{!}{
  \begin{tabular}{cc|cccc}
    \toprule
    \multicolumn{2}{c}{{\centering{Method's Variants}}} 
    & \multicolumn{2}{|c}{\centering{PASCAL VOC}}
    & \multicolumn{2}{c}{\centering{COCO Object}} \\
    \midrule
    {CSC}& {$\alpha$}  
    & mIOU & mAcc 
    & mIOU & mAcc \\
    \midrule
    \midrule
    1   & 0     & 61.9	&\underline{79.4}	&35.0	&\underline{58.2 }\\
    1   & 0.1   & 62.2	&\underline{79.4}	&35.3	&57.9 \\
    \textbf{1}   & \textbf{1 }    & \textbf{63.4}	&79.2	&\textbf{36.0}	&\textbf{58.8} \\
    1   & 10    & \underline{62.4}	&77.0	&\underline{35.8}&56.2 \\
    0   & 1     & 61.9	&\textbf{79.5}	&35.7	&55.8 \\
    \bottomrule
   \end{tabular}}
\end{table}
% anchor
\begin{table}[t]
\caption{  Ablating softness of anchor selection in InvSeg.}\label{tab:abla_anchor}
\centering
\vspace{-8pt}
\setlength{\tabcolsep}{9pt}
\resizebox{0.48\textwidth}{!}{
  \begin{tabular}{c|cccc}
    \toprule
    \multicolumn{1}{c}{{\centering{Method's Variants}}} 
    & \multicolumn{2}{|c}{\centering{PASCAL VOC}}
    & \multicolumn{2}{c}{\centering{COCO Object}} \\
    \midrule
    {scale}
    & mIOU & mAcc 
    & mIOU & mAcc \\
    \midrule
    \midrule
    2   & 63.1	& 78.4 &	35.9 &	57.3\\
    \textbf{4}     & \textbf{63.4}	& \textbf{79.2} &	\textbf{36.0 }&	{58.8}\\
    8    & 63.0	&\textbf{ 79.2} &	35.8 &	\underline{59.8}\\
    16   & 62.7	& 79.0 &	35.5 &	\textbf{59.9}\\
    % 4   & 0.7  & \textbf{63.4}	& 79.1 &	35.9 &	58.4\\
    % 4   & 0.8  & \textbf{63.4}	& 79.1 &	\textbf{36.0} &	58.6\\
    % 4   & 0.95 & \textbf{63.4}	& \textbf{79.2} &	\textbf{36.0} &	{58.8}\\
    \bottomrule
   \end{tabular}}
\end{table}

% text input
\begin{table}[ht]
\caption{ Ablating different text initialization strategies.}\label{tab:text_init}
\centering
\vspace{-8pt}
\setlength{\tabcolsep}{5pt}
\resizebox{0.48\textwidth}{!}{
  \begin{tabular}{c|cccc}
    \toprule
    \multicolumn{1}{c}{{\centering{Method's Variants}}} 
    & \multicolumn{2}{|c}{\centering{PASCAL VOC}}
    & \multicolumn{2}{c}{\centering{COCO Object}} \\
    \midrule
    {Strategies} 
    & mIOU & mAcc 
    & mIOU & mAcc \\
    \midrule
    \midrule
     \textbf{DiffSegmenter}~\cite{diffsegmenter}&\underline{ 63.4}& 	\underline{79.2}&	36.0	&\underline{58.8}\\
     BLIP~\cite{li2022blip}            & 46.3& 	53.1&	32.3	&41.6\\
     LLaVA~\cite{liu2023visual}        & 56.4& 	70.4&	33.6	&50.3\\
     COCO Caption~\cite{coco}          & -   &    -	&   \underline{38.3}	&52.8\\
     Ground Truth                      & \textbf{69.4}& 	\textbf{83.5}&	\textbf{40.3}	&\textbf{63.5}\\
    
    \bottomrule
   \end{tabular}}
\end{table}

% layers
\begin{table}[ht]
\caption{ Ablating layers to use for training in diffusion model.}\label{tab:train_layer}
\centering
\vspace{-8pt}
\setlength{\tabcolsep}{5pt}
\resizebox{0.48\textwidth}{!}{
  \begin{tabular}{cccc|cccc}
    \toprule
    \multicolumn{4}{c}{{\centering{Method's Variants}}} 
    & \multicolumn{2}{|c}{\centering{PASCAL VOC}}
    & \multicolumn{2}{c}{\centering{COCO Object}} \\
    \midrule
    Res8 & Res16 & Res32 & Res64 
    & mIOU & mAcc 
    & mIOU & mAcc \\
    \midrule
    \midrule
         \checkmark&&&&                             18.4	& 54.3 & 10.5 &	45.4   \\
         &\checkmark&&&                             \textbf{63.4}	& \textbf{79.2 }& \textbf{36.0} &	\underline{58.8}  \\
         &&\checkmark&&                             46.5	& 69.6 & 23.1 &	53.6   \\
         &&&\checkmark&                             22.9	& 57.4 & 12.3 &	45.7   \\
     \checkmark&\checkmark&&&                       54.7	& 70.4 & 28.7 &	54.3  \\
     &\checkmark&\checkmark&&                       \underline{60.7}	& \underline{77.4} & \underline{33.0} &	\textbf{59.2}  \\
     \checkmark&\checkmark&\checkmark&&             57.0	& 74.5 & 30.6 &	53.3  \\
     &\checkmark&\checkmark&\checkmark&             58.4	& 76.6 & 31.1 &	55.9  \\
     \checkmark&\checkmark&\checkmark&\checkmark&   54.8	& 72.5 & 29.3 &	56.9  \\
    \bottomrule
   \end{tabular}}
\end{table}

\subsection{Ablation Studies and Analyses}
We conduct a comprehensive analysis of various parameters and strategies on InvSeg. Our experiments are carried out on two widely-used datasets: PASCAL VOC 2012 and COCO Object, including:
loss weights,
anchor points selection,
text initialization strategies,
and diffusion model parameters.

\begin{table}[ht]
\caption{Ablating training step in diffusion model.}\label{tab:train_t}
\centering
\vspace{-8pt}
\setlength{\tabcolsep}{10pt}
\resizebox{0.48\textwidth}{!}{
   \begin{tabular}{c|cccc}
    \toprule
    \multicolumn{1}{c}{{\centering{Method's Variants}}} 
    & \multicolumn{2}{|c}{\centering{PASCAL VOC}}
    & \multicolumn{2}{c}{\centering{COCO Object}} \\
    \midrule
     {step}  
    & mIOU & mAcc 
    & mIOU & mAcc \\
    \midrule
    \midrule
    100 &61.8	&78.1	&35.5	&\underline{58.5} \\
    200 &62.0	&78.2	&35.7	&\underline{58.5} \\
    \textbf{300 }&\textbf{63.4}	&\textbf{79.2}	&\textbf{36.0}	&\textbf{58.8} \\
    500 &\underline{62.3}	&\underline{78.4}	&\underline{35.8}	&58.2 \\
    
    \bottomrule
   \end{tabular}}
\end{table}

% test  time step
\begin{table}[ht]
\caption{ Ablating time step  for inference in diffusion model.}\label{tab:test_t}
\centering
\vspace{-8pt}
\setlength{\tabcolsep}{11pt}
\resizebox{0.48\textwidth}{!}{
   \begin{tabular}{c|cccc}
    \toprule
    \multicolumn{1}{c}{{\centering{Method's Variants}}} 
    & \multicolumn{2}{|c}{\centering{PASCAL VOC}}
    & \multicolumn{2}{c}{\centering{COCO Object}} \\
    \midrule
    {step} 
    & mIOU & mAcc 
    & mIOU & mAcc \\
    \midrule
    \midrule
     25   & \underline{63.3}	&78.5	&\textbf{36.0}	&58.4 \\
     \textbf{50}   & \textbf{63.4}	&79.2	&\textbf{36.0}	&58.8 \\
     75   & \underline{63.3}	&\underline{79.6}	&\textbf{36.0}	&\underline{59.1} \\
     100  & 63.2	&\textbf{79.9}	&35.9	&\textbf{59.5} \\
     % 150  & 62.9	&77.6	&35.8	&57.1\\
     200  & 62.5	&78.1	&35.5	&58.5 \\
     300  & 60.7	&78.5	&34.3	&58.1 \\
    
    \bottomrule
   \end{tabular}}
\end{table}

\noindent \textbf{Loss weights.} To verify the effectiveness of the objective function in prompt inversion, we conducted experiments with different loss weights. In the first row, we observe that solely performing Contrastive Soft Clustering already surpasses the current state-of-the-art DiffSegmenter by 1.8\% in mIOU on VOC. Furthermore, applying the entropy minimization constraint brings more than 1\% improvement in mIOU on both datasets, which illustrates its effectiveness in stabilizing the optimization process. These results demonstrate the effectiveness of our objective function, combining Contrastive Soft Clustering and entropy minimization to achieve superior segmentation performance across different datasets.

\noindent \textbf{Anchor points selection.}
When selecting anchors from score maps, we rescale the score maps before applying sigmoid to control the steepness of the curve, which determines how much we allow the anchor selection to be soft (having non-binary values), which would result in binary selection of anchors if the scale is infinitely large. We show that the model reaches optimal performance when scale is 4. 
When the scale is 8 or larger, the performance drops, indicating that an overly rigid or near-binary selection of anchors is suboptimal for our segmentation task. 
Conversely, when the scale is less than 4, the relatively soft selection (allowing lower probabilities) can also hinder performance. 
This demonstrates the importance of finding the right balance in anchor selection softness. % for optimal segmentation results.

\noindent \textbf{Text initialization strategies.}
We evaluated the effect of using different text initialization strategies, containing the method used in DiffSegmenter, extracting class names from captions (generated by BLIP, LLaVA or GT caption), and using ground truth class names. We observe that the performance of different strategies varies significantly. The method in DiffSegmenter achieves the best overall performance, only falling behind the one using COCO captions on the COCO Object dataset. Among the remaining methods, LLaVA shows superior results compared to BLIP, mainly due to its more detailed captions. All these methods lag behind the one using ground truth class names, indicating substantial room for improvement in obtaining candidate category names.

\noindent \textbf{Diffusion model parameters.}
We conducted ablation studies on the following key parameters of our diffusion model:
1) layers to use: using different layers to optmize the text prompt can cause different result obviuosly. We aggregate the layers with same resolution together and only evaluate different resolutions. We show that when resolution is 16, we get the best performance. Combining mutiltple resolution do not bring better results.
2)time steps for training: InvSeg only runs a single pass through the diffusion process during each iteration. different time step reveals different information in the attention map. sampling in the range of 5 to 300 get the best performance. more or less will drop.
3) time steps for inference: The model achieves its best performance when t=50, with similar results at t=25. We observe that fewer time steps retain more details of the original image, with less noise added, resulting in better segmentation results.

\noindent {\textbf{Visualization.}} We present a comparison of the final segmentation results in Fig. \ref{fig:sota1}. Compared to the diffusion baseline, InvSeg produces more accurate and complete segmentation masks.
To further illustrate the improvements, we provide a visualization of the refined attention maps $\{M\}$ for both the Diffusion Baseline and InvSeg in Fig. \ref{fig:baseline}. This comparison reveals a notable difference in how each method handles various image elements. In the Diffusion Baseline, before prompt inversion is applied, we observe that background categories (such as "grass") tend to be overshadowed or suppressed by more salient foreground categories (like "cow" or "horse"). This imbalance in attention can lead to less accurate segmentation, particularly for less prominent image elements.
InvSeg, on the other hand, demonstrates a more balanced and comprehensive attention distribution across both foreground and background elements. This improved attention mechanism contributes significantly to the enhanced accuracy and completeness of the segmentation masks produced.

\section{Conclusion}
We highlight the importance of customizing image-specific text prompts to boost the potential of generative text-to-image diffusion models to segment more diverse visual concepts. To this end, we introduce region-level prompt inversion using contrastive soft clustering, leveraging the structural information embedded in pretrained diffusion models. To the best of our knowledge, this is the first unsupervised region-level prompt inversion approach. 
The proposed InvSeg is able learn more detailed spatial information for each visual concept within an image, leading to competitive segmentation results compared to previous unsupervised methods.

%%%%%%%%%%%%%%%%%%%%%%%%%%%%%%%%%%%%%%%%%%%%%%%%%%%%%%%%%%%%%%%%%%%

% partially supported by Adobe, Veritone, NSFC (62372014), CSC, and Queen Mary University of London’s Apocrita HPC facility from QMUL RESEARCH-IT.

\section{Acknowledgments}
This work was supported by Veritone, the China Scholarship Council and Queen Mary University of London’s Apocrita HPC facility from QMUL RESEARCH-IT.

\bibliography{main}
\end{document}

\maketitle

 % Ablation study
\subsection{Discussion on Other Diffusion-based Segmentors}
Several stable diffusion-based segmentation methods are not included in the main table (Tab. 1 in the main paper). 
These approaches employ diverse techniques, such as pseudo label generation, with varying computational costs, resulting in significant performance differences. We present these methods separately in Tab.~\ref{tab:sota1_more} to ensure a fair comparison among the algorithms in the main table, detailing the computational costs and corresponding performance metrics of these Stable Diffusion-based segmentors.
Our analysis reveals that the first three methods involve different amounts of real data, showing obvious superiority compared to the rest that do not rely on real data. OVAM~\cite{marcos2024open} demonstrates a noticeable performance drop when not involving real data, performing worse than our method. While OVDiff~\cite{ovdiff} requires extra time to generate data and shows superior performance on VOC and Context datasets, it still underperforms compared to our method on the COCO dataset.
\begin{table*}[ht]
\caption{ Comparison of cost and performance of existing diffusion-based segmentation  methods. $*$ notes that the result is from Mask2Former~\cite{mask2former}. All other results are from original paper or offical code source. Note that additional time required for generating pseudo data is not included in this table due to lack of available information.}\label{tab:sota1_more}
\centering
\vspace{-8pt}
\resizebox{0.99\textwidth}{!}{
  \begin{tabular}{l|cc|ccc}
    \toprule
    {\multirow{2}{*}{\centering{Methods}}} & 
  {\multirow{2}{*}{\centering{dataset (size)}}} &
  {\multirow{2}{*}{\centering{GPU time}}} 
    % \multicolumn{2}{c|}{Training dataset}
 % & 
 %    \multicolumn{2}{c|}{GPU x Time}
 % & 
 &
\multicolumn{3}{c}{mIOU}
\\ \cline{4-6}  
    & & 
     
     % & {\multirow{2}{*}{\centering{FPS}}}
     % & {\multirow{2}{*}{\centering{Memory}}}
     &
      PASCAL VOC &PASCAL Context & COCO Context
 \\
    \midrule
    \midrule

ODISE~\cite{odise} & COCO (118k) & 6d (32*V100)                               & 84.6 & 57.3 & 65.2 \\
% DiffuMask~\cite{wu2023diffumask} & Diffusion & Synthetic (60.0k)                           & 70.6 & - & - \\
%DiffSeg &  &  
DatasetDM~\cite{wu2023datasetdm} & VOC(100)+Synthetic (40.0k) & 20h(V100)           & 78.5 & - & - \\
OVAM~\cite{marcos2024open}& VOC(11.5k) +Synthetic (13.5k) & $<$1min/class (A40)   & 83.8 & - & - \\
OVAM~\cite{marcos2024open} & Synthetic (13.5k) & $<$1min/class (A40)                & 53.5 & - & - \\
OVDiff~\cite{ovdiff} & Synthetic (64/class) & 190h(A40)                            & 67.1 & 30.1 & 34.8 \\
% EMERDIFF (w. TCL) & SD v1.4 & None & 0 & - & - & - & 35.4 & - 
% DiffSegmenter~\cite{diffsegmenter} & SD v1.5 & None & 0                                        & 60.1 & 27.5 & 37.9 \\
InvSeg &  None & 0                                                                      & 63.4 & 27.8 & 36.0 \\
    \bottomrule
  \end{tabular}}
\end{table*}

\subsection{More Visualization}
\begin{figure*}[ht]
\centering
 \includegraphics[width=0.67\textwidth]{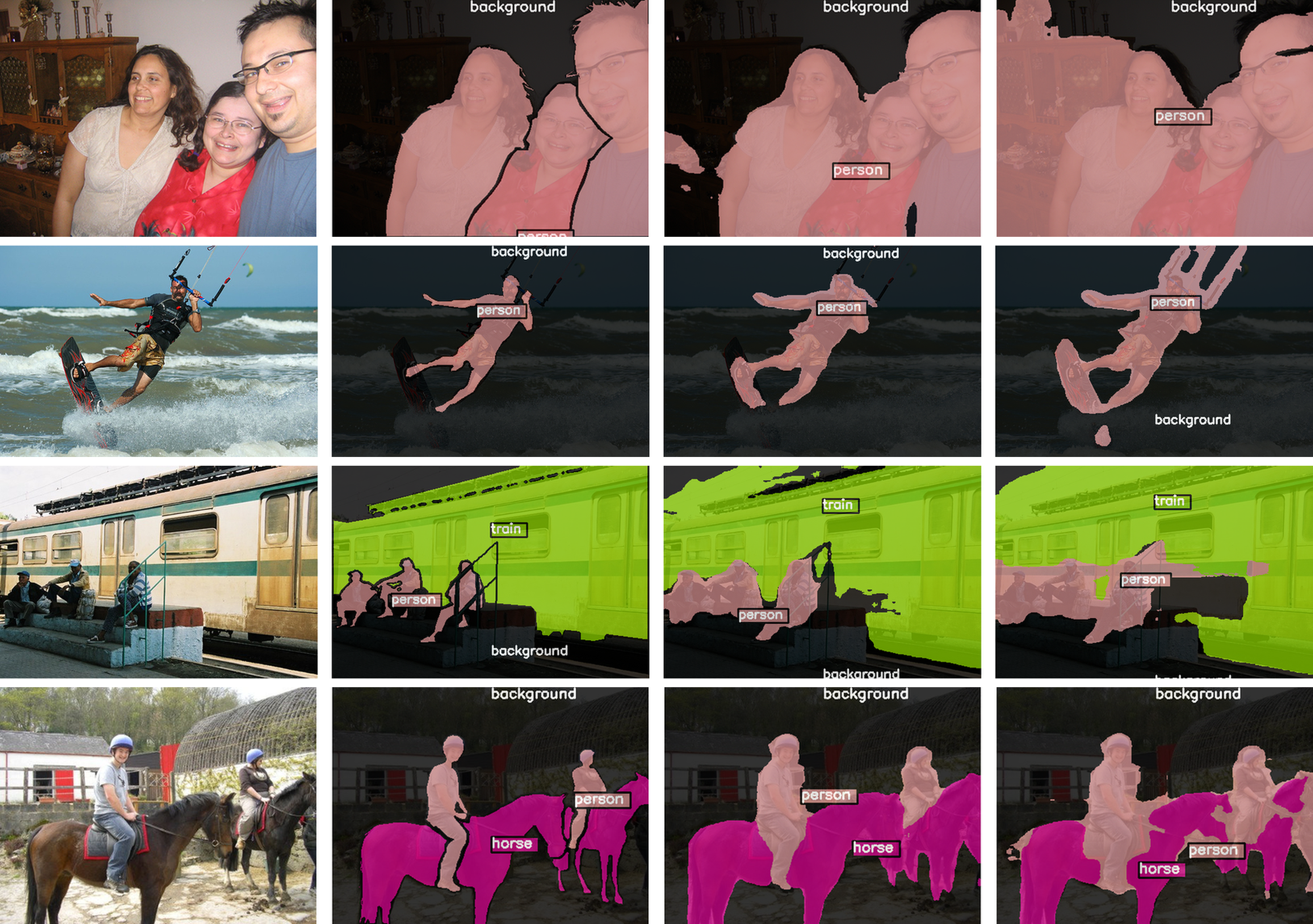}   
  \caption{Examples of Segmentation on VOC. For each image group of four, from left to right is input, GT, InvSeg, Diffusion baseline.  }
  \label{fig:sota_voc}
\end{figure*}

\begin{figure*}[ht]
\centering
 \includegraphics[width=0.67\textwidth]{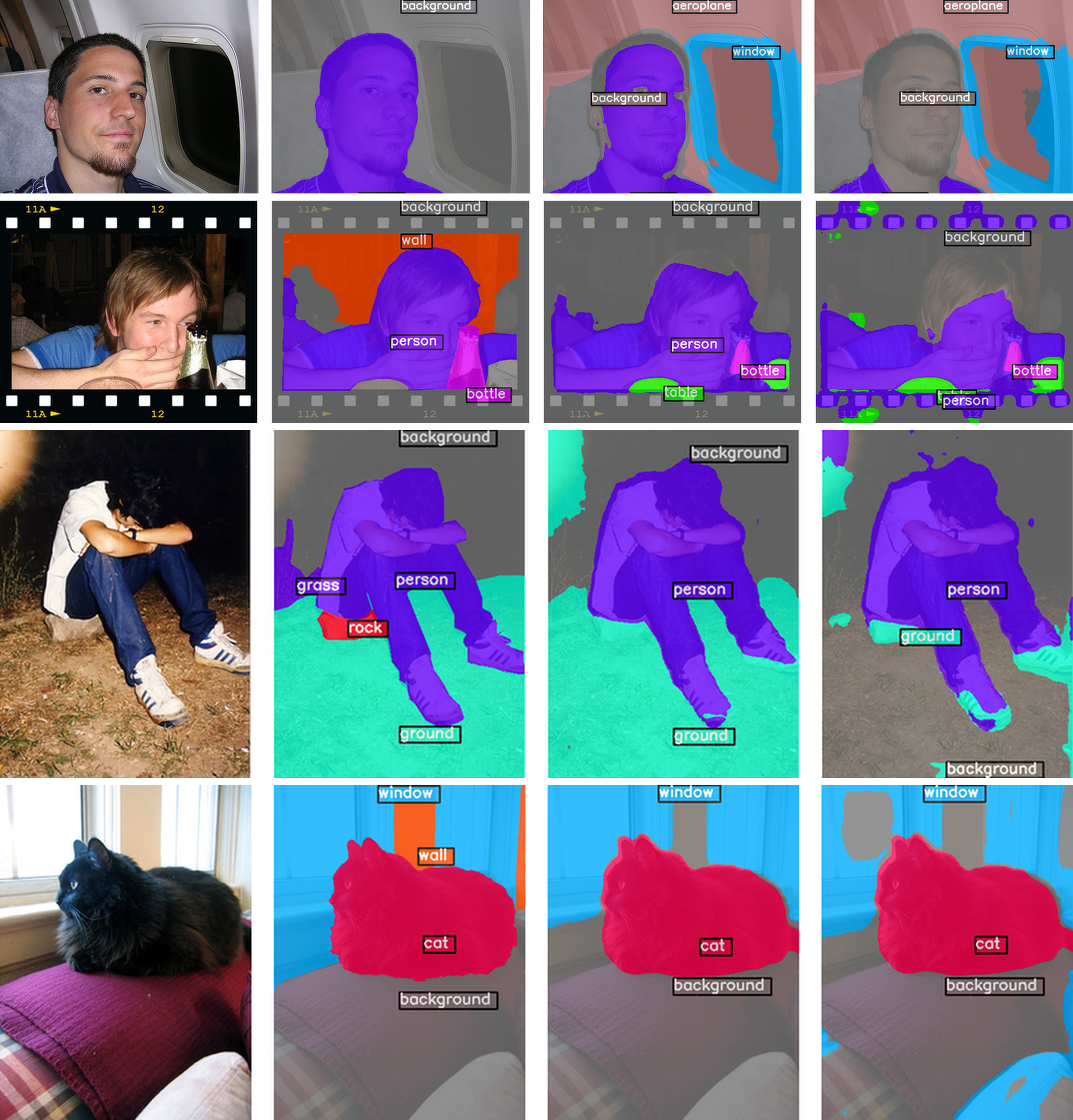}   
  \caption{Examples of Segmentation on Context. }
  \label{fig:sota_context}
\end{figure*}

\begin{figure*}[ht]
\centering
 \includegraphics[width=0.67\textwidth]{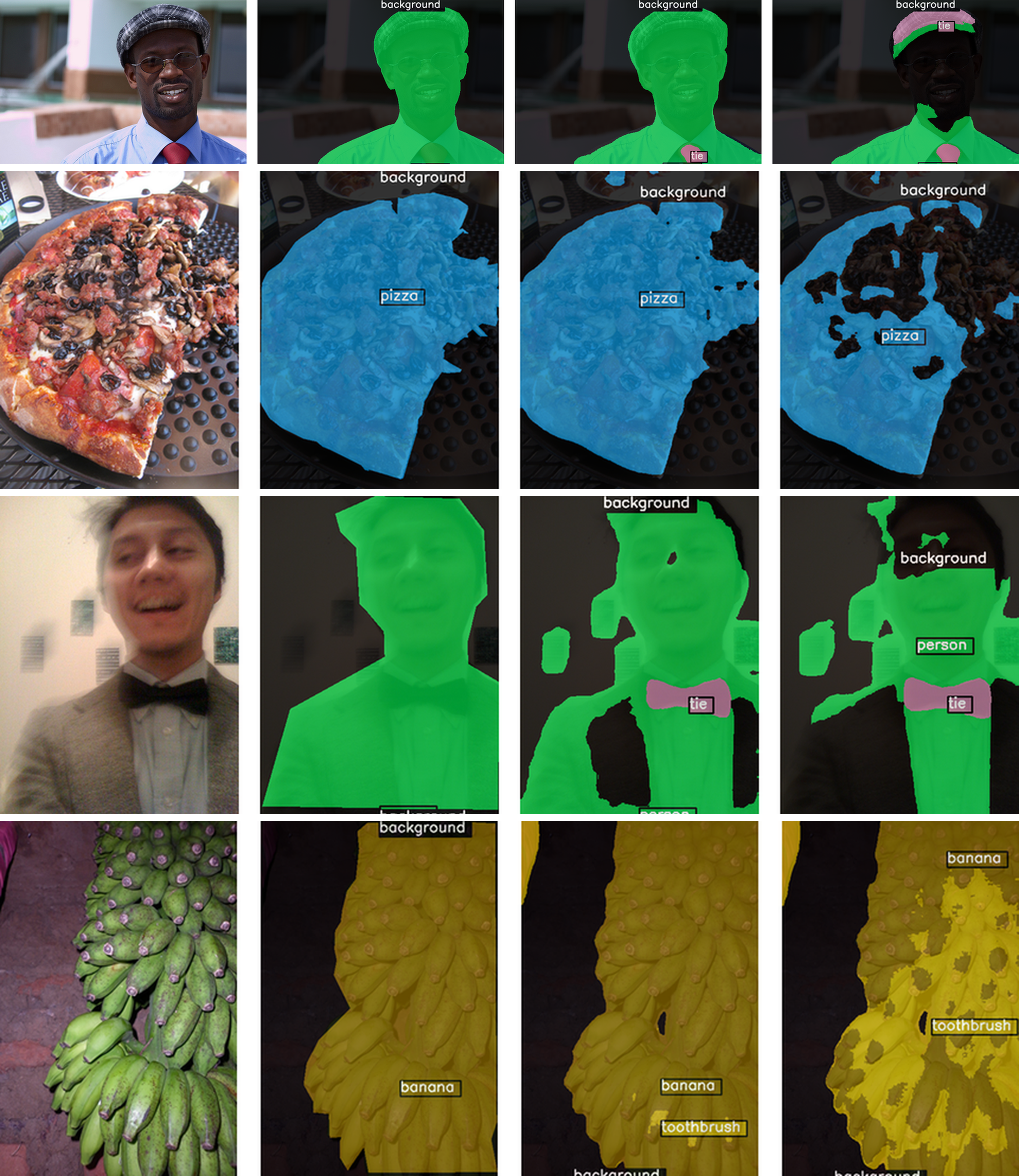}   
  \caption{Examples of Segmentation on COCO.  }
  \label{fig:sota_coco}
\end{figure*}

We present more Visualization results on three segmentation datasets PASCAL VOC 2012~\cite{everingham2010pascal}, PASCAL Context~\cite{pascal0context} and COCO Object~\cite{coco} for OVSS task (Fig.~\ref{fig:sota_voc}, Fig.~\ref{fig:sota_context}, Fig.~\ref{fig:sota_coco}).

\subsection{Limitations}
One limitation of InvSeg is that its performance lags behind approaches such as DiffSegmenter, which leverage Vision-Language Models (VLM) to produce detailed background descriptions for each category. InvSeg, on the other hand, uses only the candidate categories from the target dataset as its entire vocabulary, generating all segmentation masks simultaneously, which is relatively neat.
Furthermore, similar to previous methods that using attention map from diffusion models as segmentation mask, the resolution $64\times64$ is limited. One possible solution is to use pretrained segmentation models~\cite{kirillov2023segment} for further refinement.

\bibliography{main}